\pdfoutput=1

\documentclass[11pt]{article}

\usepackage{acl}

\usepackage{times}
\usepackage{latexsym}

\usepackage[T1]{fontenc}

\usepackage[utf8]{inputenc}

\usepackage{microtype}

\usepackage{pgfplots}
\pgfplotsset{compat=newest}
\usepgfplotslibrary{groupplots}
\usepgfplotslibrary{dateplot}

\usepackage{pdflscape}

\usepackage{graphicx}
\usepackage{caption}
\usepackage{subcaption}
\usepackage{float}

\makeatletter
\newcommand*{\centerfloat}{%
  \parindent \z@
  \leftskip \z@ \@plus 1fil \@minus \marginparwidth
  \rightskip \leftskip
  \parfillskip \z@skip}
\makeatother

\title{Towards BERT-based Automatic ICD Coding: \\ Limitations and Opportunities}

 \author{Dami\'an Pascual $\;$ \textbf{Sandro Luck} $\;$ \textbf{Roger Wattenhofer} \\
 ETH Zurich, Switzerland \\
 \texttt{\{dpascual,wattenhofer\}@ethz.ch, sluck@student.ethz.ch}
 }

\begin{document}
\maketitle
\begin{abstract}
Automatic ICD coding is the task of assigning codes from the International Classification of Diseases (ICD) to medical notes. These codes describe the state of the patient and have multiple applications, e.g., computer-assisted diagnosis or epidemiological studies. ICD coding is a challenging task due to the complexity and length of medical notes.
Unlike the general trend in language processing, no transformer model has been reported to reach high performance on this task. Here, we investigate in detail ICD coding using PubMedBERT, a state-of-the-art transformer model for biomedical language understanding. We 
find that the difficulty of fine-tuning the model on long pieces of text is the main limitation for BERT-based models on ICD coding. We run extensive experiments and show that despite the gap with current state-of-the-art, pretrained transformers can reach competitive performance using relatively small portions of text. We point at better methods to aggregate information from long texts as the main need for improving BERT-based ICD coding.
\end{abstract}

\section{Introduction}

During patient stays in medical institutions, clinicians generate text notes that record the state of the patient as well as the diagnoses and the treatments administered. 
Typically, a code from the International Classification of Diseases (ICD) is assigned to these clinical notes, in order to provide standardized information about the patient condition.
ICD codes are used for different purposes, such as billing, computer-assisted diagnosis or epidemiological studies~\cite{choi2016doctor, denny2010phewas, avati2018improving}. Assigning ICD codes to medical notes is usually done manually by clinicians. This is an error-prone and time-consuming procedure and therefore, automatic solutions have been studied for over two decades~\cite{larkey1996combining, de1998hierarchical}.

However, automatic ICD code assignment proves challenging for multiple reasons.
First, there exists a very large number of ICD codes (~$17.000$) and each clinical report may have associated more than one code. To deal with this large multi-label classification problem, it is common to reduce the number of codes to those that appear most frequently~\cite{mullenbach2018explainable}. Second, medical text usually lacks structure, includes irrelevant passages, as well as abbreviations, misspellings, numbers and a very specific vocabulary. On top of that, medical notes are long, which makes it difficult for automatic coding models to draw relations between different sections of the reports.

Current state-of-the-art methods for automatic ICD coding from medical notes are based on deep learning~\cite{wang2018joint, mullenbach2018explainable, vulabel}. These methods use different configurations of convolutional (CNN) and recurrent (RNN) neural networks as well as attention modules\cite{bahdanau2014neural}. This stands in contrast to most areas of natural language processing (NLP), where models based on the transformer architecture~\cite{vaswani2017attention} dominate the state-of-the-art~\cite{wang2019superglue}. One of the main strengths of transformer models is their ability to deal with long range dependencies. This is a desirable property in ICD coding, where an understanding of different parts of the document may be necessary to assign a code.
The lack of transformer models for ICD coding is surprising, especially since there already exist BERT-based models~\cite{devlin2019bert} (a type of bidirectional transformer) that are trained on medical text data~\cite{lee2020biobert, alsentzer2019publicly, pubmedbert}. These models have achieved state-of-the-art performance on other tasks such as named entity recognition or question answering on medical documents~\cite{pubmedbert}. 

On the other hand, the complexity of transformers scales quadratically with the length of their input, which restricts the maximum number of words that they can process at once. This limitation may be critical in ICD coding, since clinical notes usually exceed this maximum input length. In this work, we investigate in detail BERT-based ICD coding, and explore different strategies to overcome the constraint on the input length by using an encoder-decoder architecture. We use the MIMIC-III dataset~\cite{johnson2016mimic}, a big and widely used dataset for the ICD coding task, in order that our results are directly comparable to other existing methods~\cite{wang2018joint, mullenbach2018explainable, vulabel}. By exposing the limitations and benefits of BERT-based models on this task our work sets a solid basis for further research on automatic ICD coding systems.

\section{Related Work}

Automatic ICD coding has been an active area of research for over two decades. Already \citet{larkey1996combining} and \citet{de1998hierarchical} proposed different strategies to extract features from medical documents in order to build classifiers for automatically assigning ICD codes to medical notes. More recently, \citet{perotte2014diagnosis} proposed a multi-level Support Vector Machine (SVM) model to predict ICD codes from the MIMIC-II dataset~\cite{saeed2011multiparameter}, the precursor of the MIMIC-III dataset~\cite{johnson2016mimic} that we consider in this work. Similarly, \citet{scheurwegs2017selecting} presented a method to extract features from structured and unstructured text and evaluated it on the MIMIC-III dataset.

In the last years, the state-of-the-art of automatic ICD coding has been dominated by deep learning models. \citet{shi2017towards} proposed an LSTM model that operates at the character-level combined with an attention mechanism~\cite{bahdanau2014neural}. \citet{wang2018joint} proposed an embedding model based on GloVE embeddings~\cite{pennington2014glove} that maps text and labels to the same space, where predictions are made using the cosine similarity. \citet{mullenbach2018explainable} proposed a model that combined convolutions with a per-label attention mechanism. This model was further improved by \citet{xie2019ehr} and \citet{li2020icd}. \citet{vulabel}, proposed a label-attention model that reached the current best performance for ICD coding on the MIMIC-III dataset. All of these works represent only a portion of the research carried out in this field~\cite{karimi2017automatic, baumel2018multi, song2020generalized, prakash2017condensed, cao2020hypercore}.

Since the appearance of the Transformer model~\cite{vaswani2017attention}, transformer-based architectures~\cite{brown2020language, lewis2020bart, raffel2019exploring} have become state-of-the-art in almost every area of Natural Language Processing~\cite{wang2018glue, wang2019superglue} thanks to their ability to handle long range dependencies. BERT~\cite{devlin2019bert}, a bidirectional transformer, is of particular importance since it is the basis of many other language understanding models.
Nonetheless, given the specific characteristics of medical text, e.g., specialized vocabulary, models pretrained on generic language, like BERT, do not reach high performance on biomedical language understanding tasks.
Therefore, specialized models, such as BioBERT~\cite{lee2020biobert} or ClinicalBERT~\cite{alsentzer2019publicly}, pretrained on medical text have been proposed. In particular, the recent PubMedBERT model~\cite{pubmedbert} is the state-of-the-art in the BLURB benchmark~\cite{pubmedbert}, a benchmark for biomedical language understanding which includes the following tasks: named entity recognition, question answering, document classification, relation extraction, sentence similarity and evidence-based medical information extraction. 
Despite its prominence in medical language understanding, automatic ICD coding escapes the set of tasks where BERT-based models excel. 
To the best of our knowledge, no BERT-based model has been proposed yet that reaches competitive performance on ICD coding on the MIMIC-III dataset. In this work, we investigate in detail BERT-based ICD coding and identify existing limitations and opportunities.

\section{Background}~\label{sec:bckgrnd}

In this section we present the BERT model used in our experiments as well as the evaluation metrics.

\subsection{PubMedBERT}

PubMedBERT~\cite{pubmedbert} is a transformer model with the same architecture as BERT-base~\cite{devlin2019bert}, i.e., it has 12 transformer layers, $100$ million parameters and it outputs vector representations of $768$ elements. PubMedBERT is trained from scratch on PubMed text, on a dataset of $3.1$ billion words (21 GB). Furthermore, PubMedBERT has not been pretrained on the MIMIC datasets as ClinicalBERT~\cite{alsentzer2019publicly} or BlueBERT~\cite{peng2019transfer}, and therefore, we can evaluate it on MIMIC-III without information leakage from the test set. We choose this model among the existing ones because it is currently the state-of-the-art in biomedical understanding tasks as measured by the BLURB benchmark\footnote{\url{https://microsoft.github.io/BLURB/leaderboard.html}}. We use the implementation from HuggingFace~\cite{wolf2019huggingface}.

\subsection{Evaluation Metrics}

Following previous work~\cite{wang2018joint, mullenbach2018explainable, vulabel}, we report the results of our experiments using macro- and micro-averaged AUC (Area Under the ROC Curve). In a multi-class classification problem, the macro-average computes the metric (AUC in our case) for each class independently and then averages it across classes. This gives the same weight to all classes regardless of possible imbalances in the data. Micro-averaging, on the other hand, computes the average score over all samples, giving the same weight to each sample rather than to each class.

\section{Dataset}

In this work, we use the widely-used MIMIC-III dataset~\cite{johnson2016mimic}. This dataset contains medical information in various forms, however, as in previous studies~\cite{wang2018joint, mullenbach2018explainable, vulabel}, we consider exclusively the discharge summaries for ICD coding. Discharge summaries are medical notes created by doctors at the end of a stay in a medical facility and contain all the information about the stay. In the MIMIC-III dataset, the length of the discharge summaries after tokenization ranges from $78$ to $18,429$ tokens with a mean length of $2,740$ tokens and a median of $2,500$. Each of these discharge summaries has associated to it one or more ICD codes from the ICD-9 taxonomy, with an average of $13.15$ ICD codes per summary. Therefore, ICD coding is a multi-label classification task. 

The MIMIC-III dataset consists of $52,722$ discharge summaries with a total of $8,921$ unique ICD codes. However, most of the codes are very infrequent, and therefore, existing work~\cite{wang2018joint, mullenbach2018explainable, vulabel} narrows down the task to finding only the $50$ most frequent ICD codes. 
We follow this strategy and use the reduced  dataset, sometimes referred to as MIMIC-III-50. This dataset consists of a training set of $8,067$ samples, a validation set of $1,574$ samples and a test set of $1,730$ samples.
This data split is aligned with previous work, and thus, our results are directly comparable to those in the existing literature.

\subsection{Pre-processing}

We pre-process the discharge summaries from the MIMIC-III dataset following the method proposed by \citet{mullenbach2018explainable}, which is also used by other recent work~\cite{vulabel}. This way, we convert all the text to lower case and we remove all numbers. However, we do not remove infrequent words as in \cite{mullenbach2018explainable} since BERT uses WordPiece for tokenizing and hence, it does not suffer from out-of-vocabulary terms.

\section{Model}

Discharge summaries are longer than the maximum length accepted by PubMedBERT such that it fits in the memory of a modern GPU and thus, we need to split the summaries into pieces of text. In order to process more than one piece of text per summary we adopt an encoder-decoder structure, where the encoder and the decoder are trained separately. This way, the encoder is the BERT model that maps the different pieces of text to vector representations. These vector representations are then combined and decoded into ICD codes by the decoder, which can be any kind of model. 

\subsection{Encoder}\label{sec:enc}

We use PubMedBERT as the encoder of our model, as described in Section~\ref{sec:bckgrnd}. We run our experiments on TITAN RTX GPUs with 24 GB of memory, where we can fit PubMedBERT with a maximum sequence length of $512$ tokens.\footnote{Note that even if we could fit sequences of $1024$ or $2048$ tokens, they would still be shorter than  the mean and median sequence length of the summaries.} 
We devise five different strategies to split the text of the discharge summaries:

\begin{itemize}
    \item \emph{Front}: First $512$ tokens of the summary.
    \item \emph{Back}: Last $512$ tokens of the summary.
    \item \emph{Mixed}: First $256$ and the last $256$ tokens of the summary.
    \item \emph{All}: Split the whole discharge summary into consecutive chunks of $512$ tokens; since summaries are of different length, each summary is split in a different number of chunks with the last chunk being possibly shorter.
    \item \emph{Paragraph}: Given that the discharge summaries consist of named paragraphs, we select the $200$ most frequent paragraphs, i.e., those that are present most often in the discharge summaries, each with a maximum length of $512$ tokens.
\end{itemize}

PubMedBERT has been pretrained on the masked language modeling task, and therefore, it can produce generic representations of the input text. To fine-tune this model for the ICD coding task without exceeding the memory constraints we can feed only one chunk of text at a time. This way, we fine-tune five different instances of the PubMedBERT model, one per splitting strategy, using a batch size of $1$ (to ensure the model fits in memory) and a learning rate of $5e^{-4}$. In each case, the model receives as input a piece of text of a maximum length of $512$ tokens and it is trained to predict the ICD codes of the corresponding discharge summary. Note that while the text of \emph{front}, \emph{back} and \emph{mixed} corresponds always to the same part of the discharge summary, when fine-tuning the model on the \emph{paragraph} and \emph{all} splits, each training example consists of only one paragraph or chunk, respectively. Therefore, there is no alignment across training examples (each training example comes from a different section of a discharge summary),
which introduces noise to the training.

\begin{figure}[t]
\centerfloat
\begin{tikzpicture}

\definecolor{color0}{rgb}{0.12156862745098,0.466666666666667,0.705882352941177}
\definecolor{color1}{rgb}{1,0.498039215686275,0.0549019607843137}
\definecolor{color2}{rgb}{0.172549019607843,0.627450980392157,0.172549019607843}
\definecolor{color3}{rgb}{0.83921568627451,0.152941176470588,0.156862745098039}
\definecolor{color4}{rgb}{0.580392156862745,0.403921568627451,0.741176470588235}
\definecolor{color5}{rgb}{0.549019607843137,0.337254901960784,0.294117647058824}

\begin{axis}[
legend cell align={left},
legend style={font=\fontsize{10}{5}\selectfont, fill opacity=0.8, draw opacity=1.0, text opacity=1, at={(0.99,0.55)}, anchor=east, draw=white!80!black},
tick align=outside,
tick pos=left,
x grid style={white!69.0196078431373!black},
xlabel={\# Epochs},
xticklabels={0,1,2,3,4,5,6},
xmin=-0.25, xmax=5.25,
xtick style={color=black},
y grid style={white!69.0196078431373!black},
ylabel={Validation Loss},
ymin=0.19785, ymax=0.31115,
ytick style={color=black},
height = {5.5cm},
width = \linewidth
]
\addplot [very thick, color0]
table {%
0 0.238
1 0.212
2 0.206
3 0.211
4 0.205
5 0.203
};
\addlegendentry{Front}
\addplot [very thick, color1]
table {%
0 0.254
1 0.235
2 0.23
3 0.234
4 0.227
5 0.224
};
\addlegendentry{Back}
\addplot [very thick, color2]
table {%
0 0.237
1 0.213
2 0.207
3 0.214
4 0.206
5 0.205
};
\addlegendentry{Mixed}
\addplot [very thick, color3]
table {%
0 0.2957
1 0.2949
2 0.2948
3 0.295
4 0.295
5 0.295
};
\addlegendentry{Paragr.}
\addplot [very thick, color4]
table {%
0 0.26099
1 0.306
2 0.2977
3 0.28516
4 0.2933
5 0.2868
};
\addlegendentry{All}
\end{axis}

\end{tikzpicture}
\caption{Validation losses for PubMED-BERT trained on different parts of the text.}
\label{fig:val_loss}
\end{figure}
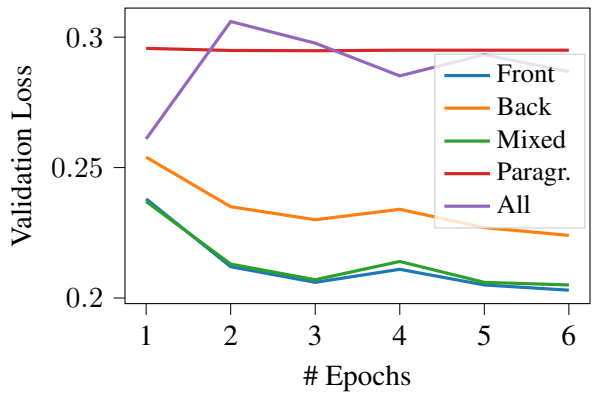

Figure \ref{fig:val_loss} depicts the validation losses after 6 epochs of training for each of the trained models. For \emph{front}, \emph{back} and \emph{mixed}, we see that the validation loss decreases quickly during the first three epochs and then, it slowly stabilizes. However, for \emph{paragraph} and \emph{all}, the validation loss stays constant, which indicates that the model is failing to learn; in other words, the lack of alignment between training samples makes the task of ICD coding too challenging for the model to learn meaningful representations of the input text.

\subsection{Decoder}\label{sec:dec}

If we consider only one part of the text at a time, PubMedBERT can directly make a prediction on the ICD codes for the corresponding summary, as done during fine-tuning. However, in order to use the information from different pieces of text, we need a decoder capable of combining the information from several encodings.
This way, the decoder receives as input one or several encoded representations (from the same discharge summary) generated by PubMedBERT during the encoding stage and outputs a vector of probabilities for the $50$ ICD codes.
For the decoder architecture, we consider a linear layer, multi-layer perceptrons (MLPs) and transformers.

In all cases, the decoders are trained 
with binary cross entropy loss with logits. We use a batch size of $32$, a learning rate of $1e^{-4}$ with linear decay for $30$ epochs and weight decay with $\lambda=1e^{-3}$. We train for a maximum of $100$ epochs with early stopping on the validation set. 

\paragraph{Linear layer} Our simplest decoder consists of a linear layer that takes as input a concatenation of the encoding vectors (of size $768$ each); when only one chunk is considered, the input is just one encoding vector. The output of this linear layer is the probability vector for the ICD codes.

\paragraph{Multi Layer Perceptron} We consider two variants of MLP-architectures, flat and parallel. In the flat architecture, the input is the concatenation of the encodings, as for the linear layer. This vector is passed through two non-linear layers, which produce intermediate representation of size $768$ and $512$ respectively, and then to a final linear layer that outputs the probabilities of the $50$ ICD codes. In the parallel architecture, each of the input encodings is processed by a different dense layer, each of which produces an output of size $768/n$, where $n$ is the number of input encodings. These intermediate representations are concatenated and passed through two additional non-linear layers, with the same sizes as in the flat architecture. 

Each of the non-linear layers includes layer normalization~\cite{ba2016layer}, PReLU activation~\cite{he2015delving}, and dropout~\cite{srivastava2014dropout} with $p=0.1$. 

\paragraph{Transformer} This decoder takes as input the encodings and treats each of them as a token of dimensionality $768$. These tokens are passed through a transformer layer with $8$ attention heads. The output of this transformer layer is of the same size as the input, i.e., a set of tokens of $768$ elements. The tokens are then concatenated and passed through an MLP of the same structure as the \emph{flat} MLP described above.

\section{Results}

We pose six research questions regarding the different strategies to encode and decode discharge summaries using a BERT-based encoder. In our experiments, we fix the random seed so that all the results are comparable.

\subsection*{How much does fine-tuning the encoder help decoding?} 

Here, we consider only the PubMedBERT models fine-tuned on \emph{front}, \emph{back} and \emph{mixed} data, since they were the only ones to learn during fine-tuning, as shown in Section~\ref{sec:enc}. 
To investigate the impact of this fine-tuning step on decoding performance, we use a simple linear layer which receives as input the concatenation of the encodings of the \emph{front}, \emph{back} and \emph{mixed} chunks. Each of these pieces of text is encoded by the PubMedBERT model trained on that piece of text, i.e., we use three different encoders. We study the difference in performance for three different training points of the encoders: not fine-tuned, fine-tuned for three epochs and fine-tuned for six epochs. The results are detailed in Table~\ref{tab:ft}.

\begin{table}[h]
\centering
\begin{tabular}{c|cc}
\textbf{Epochs} & \textbf{Macro AUC} & \textbf{Micro AUC} \\ \hline
\hline
None & 55.76 & 69.55 \\
3 & 81.47 & 86.00 \\
6 & \textbf{83.00} & \textbf{86.98} \\
\end{tabular}
\caption{Performance for different number of training epochs when combining the \emph{front}, \emph{back} and \emph{mixed} chunks with a linear decoder.}\label{tab:ft}
\end{table}

These results show that fine-tuning the encoder significantly improves the decoding performance and that the best performance is obtained after six epochs. In fact, the difference between fine-tuning for six epochs and not fine-tuning is as large as $27.24$ points for the Macro AUC score and $17.43$ points for the Micro AUC score. We observed the same pattern in all of our experiments, and therefore, in the following we will only present results with the encoder fine-tuned for six epochs, unless stated otherwise.

\subsection*{Which of the three pieces of text, \emph{front}, \emph{back} or \emph{mixed}, contains the most relevant information for ICD coding?} 

\begin{figure}
\centering
\begin{subfigure}[t]{.47\textwidth}
\centerfloat
\begin{tikzpicture}

\definecolor{color0}{rgb}{0.12156862745098,0.466666666666667,0.705882352941177}
\definecolor{color1}{rgb}{1,0.549019607843137,0}

\begin{axis}[
legend cell align={left},
legend style={font=\fontsize{10}{5}\selectfont, fill opacity=0.8, draw opacity=1, text opacity=1, draw=white!80!black},
tick align=outside,
tick pos=left,
x grid style={white!69.0196078431373!black},
xmin=-0.485, xmax=2.485,
xtick style={color=black},
xtick={0,1,2},
title={Linear},
xticklabels={Front,Mixed,Back},
y grid style={white!69.0196078431373!black},
ymin=70, ymax=93,
ytick style={color=black},
x tick label style={font=\fontsize{10}{5}\selectfont},
ylabel = {\%},
height={5.5cm},
width=\textwidth
]
\draw[draw=none,fill=color0] (axis cs:-0.35,0) rectangle (axis cs:0,76.6634000867519);
\addlegendimage{ybar,ybar legend,draw=none,fill=color0};
\addlegendentry{Mac AUC}

\draw[draw=none,fill=color0] (axis cs:0.65,0) rectangle (axis cs:1,76.0900432999731);
\draw[draw=none,fill=color0] (axis cs:1.65,0) rectangle (axis cs:2,74.2258072958014);
\draw[draw=none,fill=color1] (axis cs:2.77555756156289e-17,0) rectangle (axis cs:0.35,81.5216269853797);
\addlegendimage{ybar,ybar legend,draw=none,fill=color1};
\addlegendentry{Mic AUC}

\draw[draw=none,fill=color1] (axis cs:1,0) rectangle (axis cs:1.35,81.1833887967951);
\draw[draw=none,fill=color1] (axis cs:2,0) rectangle (axis cs:2.35,80.3539445593879);
\draw (axis cs:0.175,81.5216269853797) ++(0pt,3pt) node[
  scale=0.80,
  anchor=south,
  text=black,
  rotate=0.0
]{81.52};
\draw (axis cs:1.175,81.1833887967951) ++(0pt,3pt) node[
  scale=0.80,
  anchor=south,
  text=black,
  rotate=0.0
]{81.18};
\draw (axis cs:2.175,80.3539445593879) ++(0pt,3pt) node[
  scale=0.80,
  anchor=south,
  text=black,
  rotate=0.0
]{80.35};
\draw (axis cs:-0.175,76.6634000867519) ++(0pt,3pt) node[
  scale=0.80,
  anchor=south,
  text=black,
  rotate=0.0
]{76.66};
\draw (axis cs:0.825,76.0900432999731) ++(0pt,3pt) node[
  scale=0.80,
  anchor=south,
  text=black,
  rotate=0.0
]{76.09};
\draw (axis cs:1.825,74.2258072958014) ++(0pt,3pt) node[
  scale=0.80,
  anchor=south,
  text=black,
  rotate=0.0
]{74.23};
\end{axis}

\end{tikzpicture}
\end{subfigure}
\hfill
\begin{subfigure}[t]{.47\textwidth}
\centerfloat
\begin{tikzpicture}

\definecolor{color0}{rgb}{0.12156862745098,0.466666666666667,0.705882352941177}
\definecolor{color1}{rgb}{1,0.549019607843137,0}

\begin{axis}[
legend cell align={left},
legend style={font=\fontsize{10}{5}\selectfont, fill opacity=0.8, draw opacity=1, text opacity=1, draw=white!80!black},
tick align=outside,
tick pos=left,
x grid style={white!69.0196078431373!black},
xmin=-0.485, xmax=2.485,
xtick style={color=black},
xtick={0,1,2},
xticklabels={Front,Mixed,Back},
y grid style={white!69.0196078431373!black},
ymin=70, ymax=93,
ytick style={color=black},
x tick label style={font=\fontsize{10}{5}\selectfont},
height={5.5cm},
width=\textwidth,
title={MLP},
ylabel = {\%}
]
\draw[draw=none,fill=color0] (axis cs:-0.35,0) rectangle (axis cs:0,82.3501647586343);
\addlegendimage{ybar,ybar legend,draw=none,fill=color0};
\addlegendentry{Mac AUC}

\draw[draw=none,fill=color0] (axis cs:0.65,0) rectangle (axis cs:1,81.6903407877539);
\draw[draw=none,fill=color0] (axis cs:1.65,0) rectangle (axis cs:2,78.0371541186164);
\draw[draw=none,fill=color1] (axis cs:2.77555756156289e-17,0) rectangle (axis cs:0.35,86.9084726410081);
\addlegendimage{ybar,ybar legend,draw=none,fill=color1};
\addlegendentry{Mic AUC}

\draw[draw=none,fill=color1] (axis cs:1,0) rectangle (axis cs:1.35,86.2468525944598);
\draw[draw=none,fill=color1] (axis cs:2,0) rectangle (axis cs:2.35,83.7317704400527);
\draw (axis cs:0.175,86.9084726410081) ++(0pt,3pt) node[
  scale=0.80,
  anchor=south,
  text=black,
  rotate=0.0
]{86.91};
\draw (axis cs:1.175,86.2468525944598) ++(0pt,3pt) node[
  scale=0.80,
  anchor=south,
  text=black,
  rotate=0.0
]{86.25};
\draw (axis cs:2.175,83.7317704400527) ++(0pt,3pt) node[
  scale=0.80,
  anchor=south,
  text=black,
  rotate=0.0
]{83.73};
\draw (axis cs:-0.175,82.3501647586343) ++(0pt,3pt) node[
  scale=0.80,
  anchor=south,
  text=black,
  rotate=0.0
]{82.35};
\draw (axis cs:0.825,81.6903407877539) ++(0pt,3pt) node[
  scale=0.80,
  anchor=south,
  text=black,
  rotate=0.0
]{81.69};
\draw (axis cs:1.825,78.0371541186164) ++(0pt,3pt) node[
  scale=0.80,
  anchor=south,
  text=black,
  rotate=0.0
]{78.04};
\end{axis}

\end{tikzpicture}
\end{subfigure}%
\caption{Performance of a linear layer (top) and a non-linear MLP (bottom) on the \emph{front}, \emph{back} and \emph{mixed} encodings.}
\label{fig:fbmor}
\end{figure}
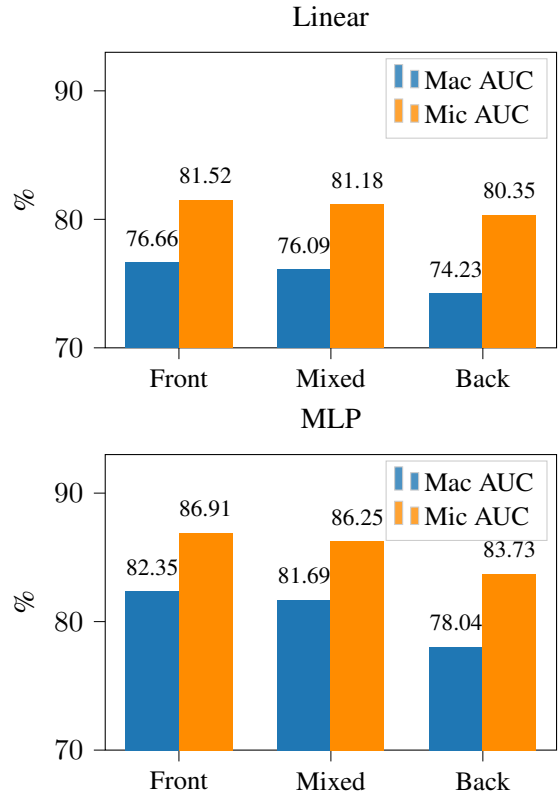

We experiment with a linear and a flat MLP decoder and apply these models to the encodings of each of the three chunks of text separately, i.e., \emph{front}, \emph{back} and \emph{mixed}. We report the results in Figure~\ref{fig:fbmor}.

We see that \emph{front}, i.e., the first $512$ tokens of the discharge summary yields the best performance, both when the decoder is a linear layer and an MLP. Although slightly inferior, the \emph{mixed} chunk produces competitive scores while when using an MLP the AUC scores are more than $3$ points lower for \emph{back} than for \emph{front}. 
Furthermore, using as decoder an MLP improves the performance significantly over using a linear layer; with the \emph{front} non-linear model performing comparably to the combination of the three chunks with a linear decoder, as reported in the previous section, Table~\ref{tab:ft}.

This naturally raises the question of whether the combination of the chunks yields an improvement. To study this, we use the same non-linear MLP architecture as in Figure~\ref{fig:fbmor} (bottom) on 1) the concatenation of the encodings of \emph{front} and \emph{back} and 2) the concatenation of the three encodings, \emph{front}, \emph{back} and \emph{mixed}. We report the results in Table~\ref{tab:fbm}.

\begin{table}[h]
\centering
\begin{tabular}{c|cc}
\textbf{Model} & \textbf{Mac. AUC} & \textbf{Mic. AUC} \\ \hline
\hline
Front-Back & 83.70 & 88.11 \\
Front-Back-Mixed & \textbf{84.42} & \textbf{88.58} \\
\end{tabular}
\caption{Performance of combining the \emph{front}, \emph{back} and \emph{mixed} chunks using a two-layer flat MLP decoder.}\label{tab:fbm}
\end{table}

These results show that combining \emph{front} and \emph{back} improves performance in comparison to using only \emph{front}. As it may be expected, adding the mixed paragraph, which contains redundant information, produces only a small improvement. Overall, the combination of the three chunks produces an improvement of $2.07$ points for Macro AUC and $1.67$ points for Micro AUC over using only \emph{front}. Given the larger input, these models have more parameters than the ones using only one of the chunks, which could partly explain the improvement, especially when adding redundant information, i.e., the \emph{mixed} chunk. This result leads us to investigate the influence of the decoder architecture.

\subsection*{How does the architecture of the decoder impact performance?}

Here, we consider flat MLP, parallel MLP and transformer decoders on the combination of \emph{front}, \emph{back} and \emph{mixed}. For each of these architectures, we evaluate three different sizes: Base, Large and X-Large, where the difference between these sizes is only the number of layers and the size of the internal representations. This way, our experiments aim at discerning whether the structure of the decoder, the number of parameters, or both, influence the performance of the ICD coding model. Table~\ref{tab:decArch} details the results of these experiments.

\begin{table}[h]
\centering
\begin{tabular}{c|cc}
\textbf{Model} & \textbf{AUC Mac.} & \textbf{AUC Mic.} \\ \hline
\hline
Flat ($1.5M$) & 84.42 & 88.58 \\
Flat L ($3M$) & 84.30 & 88.45  \\     
Flat XL ($7M$) & 84.30 & 88.47  \\     
\hline
Parallel ($1M$) & 84.45 & 88.65 \\
Parallel L ($2M$)& 84.23 & 88.48 \\ 
Parallel XL ($3M$) & 84.51 & 88.49 \\ 
\hline
Transformer ($6.5M$) & 84.30 & 88.49 \\
Transformer L ($14M$) & 84.27 & 88.45 \\ 
Transformer XL ($18M$) & 84.29 & 88.08 \\ 
\end{tabular}
\caption{Performance of different decoder architectures for the combination of \emph{front}, \emph{back} and \emph{mixed}, the number of parameters of each model is specified in parenthesis.}\label{tab:decArch}
\end{table}

None of the models considered obtains a performance significantly higher than the others, with the largest difference across Macro and Micro AUC scores being of only $0.28$ and $0.57$ points, respectively. This result is surprising since, given the complexity of the task, it could be expected that larger and more sophisticated decoders would perform better. Notwithstanding, the saturation in performance suggests that all the information available in the input of the decoder is successfully extracted by every model, regardless of its complexity. This in turn indicates that the performance of the whole encoder-decoder model is limited by the reduced amount of text that is given as input (only the beginning and the end of the discharge summaries). Therefore, we next consider providing larger portions of text from the discharge summaries as input.

\subsection*{Is dividing the discharge summaries by paragraphs a good splitting strategy?} 

By splitting the discharge summaries into paragraphs we take into account information from a larger body of text than by using the front and the back. The main disadvantage of this approach is that the encoder fails to converge during fine-tuning. Here, we test the hypothesis of whether the decoder can compensate the lack of fine-tuning of the encoder and, by leveraging the larger amount of information available, reach competitive performance. We encode the $200$ most frequent paragraphs using the PubMedBERT model fine-tuned on \emph{paragraph} data, although due to lack of convergence during fine-tuning, we observed very similar results when using the not fine-tuned version.

Since not all the discharge summaries contain the same paragraphs, there is a misalignment between samples. For this reason, here we consider only the transformer decoder; the self-attention modules of the transformer should be able to cope with the misalignment better than the other architectures. We consider the transformer decoders (Base, Large and X-Large) from the previous section. Now, the transformer decoder receives $200$ encoded representations, one per paragraph. Given this large number of input representations or tokens, we aggregate the output of the transformers by taking the mean over the representations produced for all the paragraphs\footnote{We experimented with other aggregation techniques like max pooling and large MLPs obtaining very similar results.}.

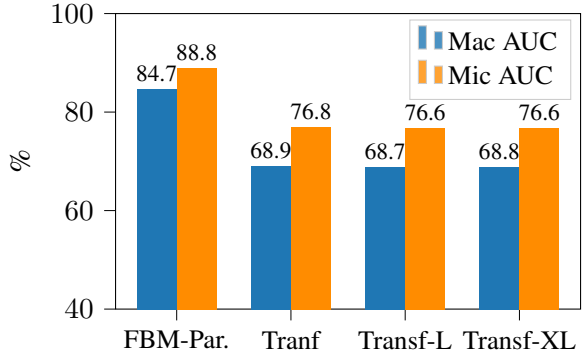
\begin{figure}[t]
\centerfloat
\begin{tikzpicture}

\definecolor{color0}{rgb}{0.12156862745098,0.466666666666667,0.705882352941177}
\definecolor{color1}{rgb}{1,0.549019607843137,0}

\begin{axis}[
legend cell align={left},
legend style={font=\fontsize{10}{5}\selectfont, fill opacity=0.8, draw opacity=1, text opacity=1, draw=white!80!black},
tick align=outside,
tick pos=left,
x grid style={white!69.0196078431373!black},
xmin=-0.535, xmax=3.535,
xtick style={color=black},
xtick={0,1,2,3},
xticklabels={FBM-Par., Tranf, Transf-L, Transf-XL},
x tick label style={font=\fontsize{10}{5}\selectfont},
y grid style={white!69.0196078431373!black},
ymin=40, ymax=100,
ytick style={color=black},
ylabel={\%},
height = {5.5cm},
width = \linewidth
]
\draw[draw=none,fill=color0] (axis cs:-0.35,0) rectangle (axis cs:0,84.6583511265549);
\addlegendimage{ybar,ybar legend,draw=none,fill=color0};
\addlegendentry{Mac AUC}

\draw[draw=none,fill=color0] (axis cs:0.65,0) rectangle (axis cs:1,68.8947344591349);
\draw[draw=none,fill=color0] (axis cs:1.65,0) rectangle (axis cs:2,68.7053474886055);
\draw[draw=none,fill=color0] (axis cs:2.65,0) rectangle (axis cs:3,68.7569278399686);
\draw[draw=none,fill=color1] (axis cs:2.77555756156289e-17,0) rectangle (axis cs:0.35,88.7734277815453);
\addlegendimage{ybar,ybar legend,draw=none,fill=color1};
\addlegendentry{Mic AUC}

\draw[draw=none,fill=color1] (axis cs:1,0) rectangle (axis cs:1.35,76.8110736282798);
\draw[draw=none,fill=color1] (axis cs:2,0) rectangle (axis cs:2.35,76.609903830871);
\draw[draw=none,fill=color1] (axis cs:3,0) rectangle (axis cs:3.35,76.6498084252723);
\draw (axis cs:0.175,88.7734277815453) ++(0pt,0pt) node[
  scale=0.80,
  anchor=south,
  text=black,
  rotate=0.0
]{88.8};
\draw (axis cs:1.175,76.8110736282798) ++(0pt,0pt) node[
  scale=0.80,
  anchor=south,
  text=black,
  rotate=0.0
]{76.8};
\draw (axis cs:2.175,76.609903830871) ++(0pt,0pt) node[
  scale=0.80,
  anchor=south,
  text=black,
  rotate=0.0
]{76.6};
\draw (axis cs:3.175,76.6498084252723) ++(0pt,0pt) node[
  scale=0.80,
  anchor=south,
  text=black,
  rotate=0.0
]{76.6};
\draw (axis cs:-0.175,84.6583511265549) ++(0pt,0pt) node[
  scale=0.80,
  anchor=south,
  text=black,
  rotate=0.0
]{84.7};
\draw (axis cs:0.825,68.8947344591349) ++(0pt,0pt) node[
  scale=0.80,
  anchor=south,
  text=black,
  rotate=0.0
]{68.9};
\draw (axis cs:1.825,68.7053474886055) ++(0pt,0pt) node[
  scale=0.80,
  anchor=south,
  text=black,
  rotate=0.0
]{68.7};
\draw (axis cs:2.825,68.7569278399686) ++(0pt,0pt) node[
  scale=0.80,
  anchor=south,
  text=black,
  rotate=0.0
]{68.8};
\end{axis}

\end{tikzpicture}
\caption{Comparison of front-back-mixed parallel (FBM-Par.) and three sizes of transformer decoders (Transf) on \emph{paragraph} data.}
\label{fig:paras_vs_fbm}
\end{figure}

In Figure~\ref{fig:paras_vs_fbm}, we compare these \emph{paragraph} decoders to the Parallel MLP model on the \emph{front}, \emph{back} and \emph{mixed} chunks from the previous section.

We see that dividing the discharge summaries into paragraphs 
greatly under-performs in comparison to using the beginning and end of the summaries encoded by fine-tuned PubMedBERT models. 
This result partly rejects the hypothesis that the decoder can benefit from a larger unstructured input. Next, we continue investigating this hypothesis by feeding the decoder with the complete discharge summaries following the \emph{all} strategy.

\subsection*{How does splitting the complete summaries in consecutive chunks perform?} 

\begin{figure}[t]
\centerfloat
\begin{tikzpicture}

\definecolor{color0}{rgb}{0.12156862745098,0.466666666666667,0.705882352941177}
\definecolor{color1}{rgb}{1,0.549019607843137,0}

\begin{axis}[
legend cell align={left},
legend style={font=\fontsize{10}{5}\selectfont, fill opacity=0.8, draw opacity=1, text opacity=1, draw=white!80!black},
tick align=outside,
tick pos=left,
x grid style={white!69.0196078431373!black},
xmin=-0.535, xmax=3.535,
xtick style={color=black},
xtick={0,1,2,3},
xticklabels={FBM-Par.,Transf,Transf-L,Transf-XL},
y grid style={white!69.0196078431373!black},
ymin=40, ymax=100,
ytick style={color=black},
x tick label style={font=\fontsize{10}{5}\selectfont},
height = {5.5cm},
ylabel={\%},
width = \linewidth
]
\draw[draw=none,fill=color0] (axis cs:-0.35,0) rectangle (axis cs:0,84.6583511265549);
\addlegendimage{ybar,ybar legend,draw=none,fill=color0};
\addlegendentry{Mac AUC}

\draw[draw=none,fill=color0] (axis cs:0.65,0) rectangle (axis cs:1,45.3183355103516);
\draw[draw=none,fill=color0] (axis cs:1.65,0) rectangle (axis cs:2,47.9585264827785);
\draw[draw=none,fill=color0] (axis cs:2.65,0) rectangle (axis cs:3,50.4989210289419);
\draw[draw=none,fill=color1] (axis cs:2.77555756156289e-17,0) rectangle (axis cs:0.35,88.7734277815453);
\addlegendimage{ybar,ybar legend,draw=none,fill=color1};
\addlegendentry{Mic AUC}

\draw[draw=none,fill=color1] (axis cs:1,0) rectangle (axis cs:1.35,64.8772926839799);
\draw[draw=none,fill=color1] (axis cs:2,0) rectangle (axis cs:2.35,68.1379378823607);
\draw[draw=none,fill=color1] (axis cs:3,0) rectangle (axis cs:3.35,68.7363235040542);
\draw (axis cs:0.175,88.7734277815453) ++(0pt,0pt) node[
  scale=0.80,
  anchor=south,
  text=black,
  rotate=0.0
]{88.8};
\draw (axis cs:1.175,64.8772926839799) ++(0pt,0pt) node[
  scale=0.80,
  anchor=south,
  text=black,
  rotate=0.0
]{64.9};
\draw (axis cs:2.175,68.1379378823607) ++(0pt,0pt) node[
  scale=0.80,
  anchor=south,
  text=black,
  rotate=0.0
]{68.1};
\draw (axis cs:3.175,68.7363235040542) ++(0pt,0pt) node[
  scale=0.80,
  anchor=south,
  text=black,
  rotate=0.0
]{68.7};
\draw (axis cs:-0.175,84.6583511265549) ++(0pt,0pt) node[
  scale=0.80,
  anchor=south,
  text=black,
  rotate=0.0
]{84.7};
\draw (axis cs:0.825,45.3183355103516) ++(0pt,0pt) node[
  scale=0.80,
  anchor=south,
  text=black,
  rotate=0.0
]{45.3};
\draw (axis cs:1.825,47.9585264827785) ++(0pt,0pt) node[
  scale=0.80,
  anchor=south,
  text=black,
  rotate=0.0
]{48.0};
\draw (axis cs:2.825,50.4989210289419) ++(0pt,0pt) node[
  scale=0.80,
  anchor=south,
  text=black,
  rotate=0.0
]{50.5};
\end{axis}

\end{tikzpicture}
\caption{Comparison of front-back-mixed parallel (FBM-Par.) and three sizes of transformer decoders (Transf) on \emph{all} data.}
\label{fig:all_vs_fbm}
\end{figure}
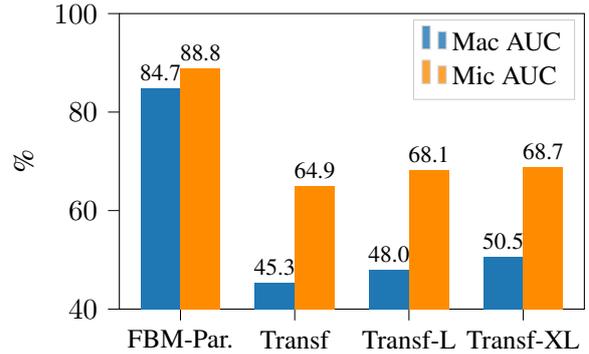

We split the whole text of each discharge summary into consecutive chunks of $512$ tokens (the last chunk of each summary may be smaller). We encode these chunks using the PubMedBERT model fine-tuned on \emph{all} data; as before, we observed very similar results with the not fine-tuned model. The encodings are then fed into the decoder. Again, the varying size of the discharge summaries produces misalignment across examples. Therefore, we consider only the transformer decoders (Base, Large and X-Large). We report the results of this experiment in Figure~\ref{fig:all_vs_fbm}.

The largest transformer model (XL) performs the best of the three models on \emph{all} data. Nevertheless, its $50.5\%$ Macro and $68.7\%$ Micro AUC scores are much lower than the results obtained by the \emph{front-back-mixed}. In fact, splitting the text into chunks of the same size performs the worst among all the methods that we have investigated. These results confirm that the decoder cannot compensate the lack of convergence during the fine-tuning of the encoder and points at the encoder as the main responsible for the model's performance.

\subsection*{How do our results compare to the state-of-the-art?}

Finally, in Table~\ref{tab:sota} we compare one of our best performing BERT-ICD models, the \emph{front-back-mixed} Parallel model, with the existing state-of-the art models for ICD coding on the MIMIC-III dataset. In particular, we compare against the condensed memory networks (C-MemNN) by \citet{prakash2017condensed}, LEAM by \citet{wang2018joint}, CAML and DR-CAML by \citet{mullenbach2018explainable}, MSATT-KG by \citet{xie2019ehr} and the Label Attention model by \citet{vulabel}. We report the performance of each model as provided in the corresponding original work. 

\begin{table}[h]
\centering
\begin{tabular}{c|cc}
\textbf{Model} & \textbf{AUC Mac.} & \textbf{AUC Mic.} \\ \hline
\hline
C-MemNN & 83.3 & -\\
LEAM & 88.1 & 91.2 \\
CAML & 87.5 & 90.9 \\
DR-CAML & 88.0 & 90.2  \\
MSATT-KG & 91.4 & 93.6 \\
Label Attention & \textbf{92.1} & \textbf{94.6} \\
\hline
BERT-ICD & 84.45 & 88.65 \\
\end{tabular}
\caption{Comparison of different state-of-the-art models for ICD coding.}\label{tab:sota}
\end{table}

We see that although our BERT-based ICD coding model is competitive with some of the state-of-the-art models, there is still a substantial gap between the best performing model from \citet{vulabel}, and our BERT-ICD model.

\section{Discussion}

Automatic ICD coding from discharge summaries using transformer models has proven to be challenging. Discharge summaries are very long documents and thus, they need to be divided into chunks in order to be entirely processed by BERT-like models.
This way, a decoder is necessary to combine the representations of each chunk, which are independently generated by the BERT encoder. We have shown that for these representations to be meaningful the encoder needs to be fine-tuned on the ICD coding task.
It is straight-forward to fine-tune a BERT encoder such as PubMedBERT using specific parts of the summary, e.g., the beginning or the end. However, in our experiments, fine-tuning PubMedBERT on excerpts extracted from different parts of the text, i.e., \emph{paragraph} and \emph{all}, prevented convergence due to the lack of alignment between samples, i.e., due to each training sample containing information from a different section of a discharge summary.
Furthermore, our results show that the decoder, regardless of its architecture, cannot compensate for lack of convergence during the fine-tuning of the encoder. 

On the other hand, our best BERT-ICD model reaches competitive performance using only $1,024$ tokens (\emph{front} and \emph{back}), which represents a significantly smaller portion of text than state-of-the-art models, based on CNNs and RNNs. 
Unlike BERT, CNN and RNN models can extract information from texts of any length without needing to split them, which allows for end-to-end training over long pieces of text. 
\citet{mullenbach2018explainable} found that the performance of their convolutional attention model benefits from longer input texts until a length of between $2,500$ and $6,500$ words, and \citet{vulabel} use up to $4,000$ words as input. 
Our model combines encodings from the beginning and the end of the discharge summary, and reaches better performance in that case than when it processes either of those portions of text alone. This supports the statement that including more text improves ICD coding. All of these results suggest that the difficulty of fine-tuning a BERT encoder on long pieces of text is the main bottleneck for performance and the reason for the existing gap with state-of-the-art models for ICD coding. 

One of the main advantages of transformer models over CNNs and RNNs is that they can handle long-range dependencies. Hence, if longer text could be fed at once into a BERT encoder it would be possible to find relationships and patterns over longer spans of text.
It is therefore likely that advances either in terms of hardware, i.e., larger GPU memories allowing for longer pieces of text to be processed at once; or in compressing BERT-like models, e.g., distillation, will progressively close the gap with the state-of-the-art, following the same trend of other areas of NLP. On top of that, we consider that the two most promising directions for future research on BERT-based ICD coding are: 1) devising strategies to fine-tune the encoder over longer spans of text, e.g., building an ensemble of models where each of them is trained on one section of the text; 2) improving the methods to aggregate encodings from different parts of the document.

Finally, to deploy automatic ICD coding systems in the real world, it is important that their decisions can be explained. Explaining transformer models is currently a field of active research, and although there exist important concerns about the interpretability of attention distributions in transformers~\cite{brunner2019identifiability, pruthi2020learning}, methods based on gradient attribution~\cite{pascual2020telling} or on attention flow~\cite{abnar2020quantifying} can provide insights on their decision-making. A BERT-based ICD coding system could directly benefit from this field of research and eventually provide explanations together with its ICD code predictions.

\section{Conclusion}

Contrary to what is common in most NLP tasks, the transformer architecture is not the state-of-the-art in assigning ICD codes to discharge summaries. In this work, we have presented a thorough study of the performance of BERT-based models on this task and we have identified the length of the discharge summaries as the main obstacle holding back their performance. Our work sets a solid foundation for further research on ICD coding and suggests that overcoming the exposed limitations of BERT-based models is likely to lead to a new state-of-the-art. Furthermore, we believe that the interpretability of ICD coding models is an interesting avenue for future work, which can benefit from a large body of existing research.

\bibliography{anthology,custom}
\bibliographystyle{acl_natbib}

\end{document}